\documentclass{article}

    \usepackage[preprint]{neurips_2019}

\usepackage[utf8]{inputenc} 
\usepackage[T1]{fontenc}    
\usepackage{url}            
\usepackage{booktabs}       
\usepackage{amsfonts}       
\usepackage{nicefrac}       
\usepackage{microtype}      
\usepackage{graphicx}       
\usepackage{color}
\usepackage[table,xcdraw]{xcolor}
\usepackage{natbib}
\usepackage{amssymb,amsmath,amsthm,amsfonts,dsfont}

\def\Z{\mathcal{Z}}
\def\R{\mathbb{R}}
\def\loss{\ell}
\def\our{\mathrm{constrainedIBP}}
\def\IBP{\mathrm{IBP}}

\title{Fast and Stable Interval Bounds Propagation for Training Verifiably Robust Models}

\author{
 Pawe\l{} Morawiecki\\
 Institute of Computer Science\\Polish Academy of Sciences\\
 Jana Kazimierza 5, Warsaw, Poland\\
  \texttt{pawel.morawiecki@gmail.com} \\
 \And
 Przemys\l{}aw Spurek\\
 Mathematics and Computer Science\\
 Jagiellonian University\\
 \L{}ojasiewicza 6, Krak\'ow, Poland\\
  \texttt{przemyslaw.spurek@uj.edu.pl} \\
 \And
  Marek \'Smieja\\
 Mathematics and Computer Science\\
 Jagiellonian University\\
 \L{}ojasiewicza 6, Krak\'ow, Poland\\
  \texttt{marek.smieja@uj.edu.pl} \\
  \And
  Jacek Tabor\\
 Mathematics and Computer Science\\
 Jagiellonian University\\
 \L{}ojasiewicza 6, Krak\'ow, Poland\\
  \texttt{jacek.tabor@uj.edu.pl} \\
}

\begin{document}

\maketitle

\begin{abstract}
   We present an efficient technique, which allows to train classification networks which are verifiably robust against norm-bounded adversarial attacks. This framework is built upon the work of Gowal et al., who applies the interval arithmetic to bound the activations at each layer and keeps the prediction invariant to the input perturbation. While that method is faster than competitive approaches, it requires careful tuning of hyper-parameters and a large number of epochs to converge. To speed up and stabilize training, we supply the cost function with an additional term, which encourages the model to keep the interval bounds at hidden layers small. Experimental results demonstrate that we can achieve comparable (or even better) results using a smaller number of training iterations, in a more stable fashion. Moreover, the proposed model is not so sensitive to the exact specification of the training process, which makes it easier to use by practitioners.
\end{abstract}

\section{Introduction}

Deep learning models achieve impressive performance in computer vision \cite{krizhevsky2012imagenet}, natural language processing \cite{hinton2012deep}, and many other domains. Although neural networks are able to outperform humans on various machine learning tasks, they are also vulnerable to adversarial examples \cite{szegedy2013intriguing}. In particular, a slightly modified input can fool the neural model and change its prediction. This is a serious problem, which limits the use of NNs in many areas, such as autonomous cars \cite{sitawarin2018darts} or malware detection \cite{grosse2017adversarial}, where security is a priority. In recent years, a lot of effort has been put on understanding deep learning models and making them more robust \cite{salman2019convex, mirman2019provable}.

Adversarial attacks rely on creating such input data points, which are visually indistinguishable from `normal' examples, but drastically change the prediction of the model \cite{goodfellow2014explaining}. One remedy is to construct adversarial examples and add them to the training set \cite{madry2017towards}. While such models become robust to many adversarial attacks, there are no guarantees that another adversarial scheme exists. To formally verify the robustness of the model against norm-bounded perturbations, one can find the outer bound on the so-called `adversarial polytope' \cite{wong2017provable}. These techniques give loose bounds on the output activations, but guarantee that no adversary within a given norm can change the class label. Unfortunately, most of these techniques are computationally demanding and do not scale well to large networks, which makes them difficult to use in practice.



\begin{figure}[t] 
\center{\includegraphics[width=\textwidth] {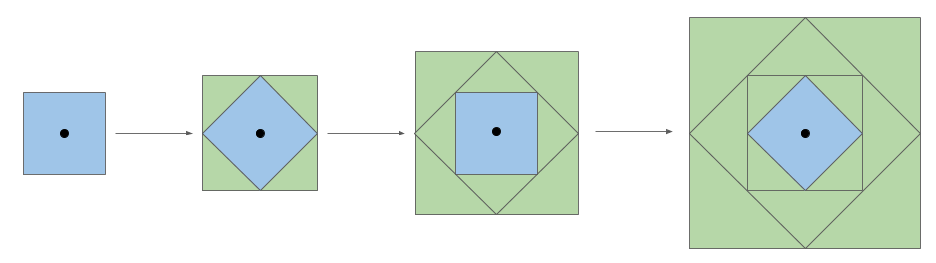}} 
\caption{The wrapping effect. A datapoint (black dot) with its initial adversarial polytope (blue square) is transformed by rotation by $45$ degrees. After every transformation a new adversarial polytope is formed (outer green square), $\sqrt{2}$ bigger than the previous one.} 
\label{fig:wrapping_effect}
\end{figure}

In this paper, we consider the framework of interval bounds propagation (IBP) proposed by Gowal et al. \cite{gowal2018effectiveness} for constructing provably robust classifiers. IBP uses the interval arithmetic to minimize the upper bound on the maximum difference between any pair of logits when the input is perturbed within the norm-bounded ball. Direct application of interval arithmetic in a layer-wise fashion leads to the well-known \textit{wrapping effect} \cite{moore1979methods}, because bounds are reshaped to be axis-aligned with bounding boxes that always encompass the adversarial polytope, see Figure \ref{fig:wrapping_effect}. To overcome this limitation, the authors starts from a typical classification loss to pretrain the network and gradually increases the importance of adversarial loss together with increasing the size of the input perturbation. Unfortunately, too sudden change of these trade-off factors results in the lack of convergence, which makes the training process cumbersome and time consuming.

\begin{figure}[t] 
\center{\includegraphics[width=\textwidth] {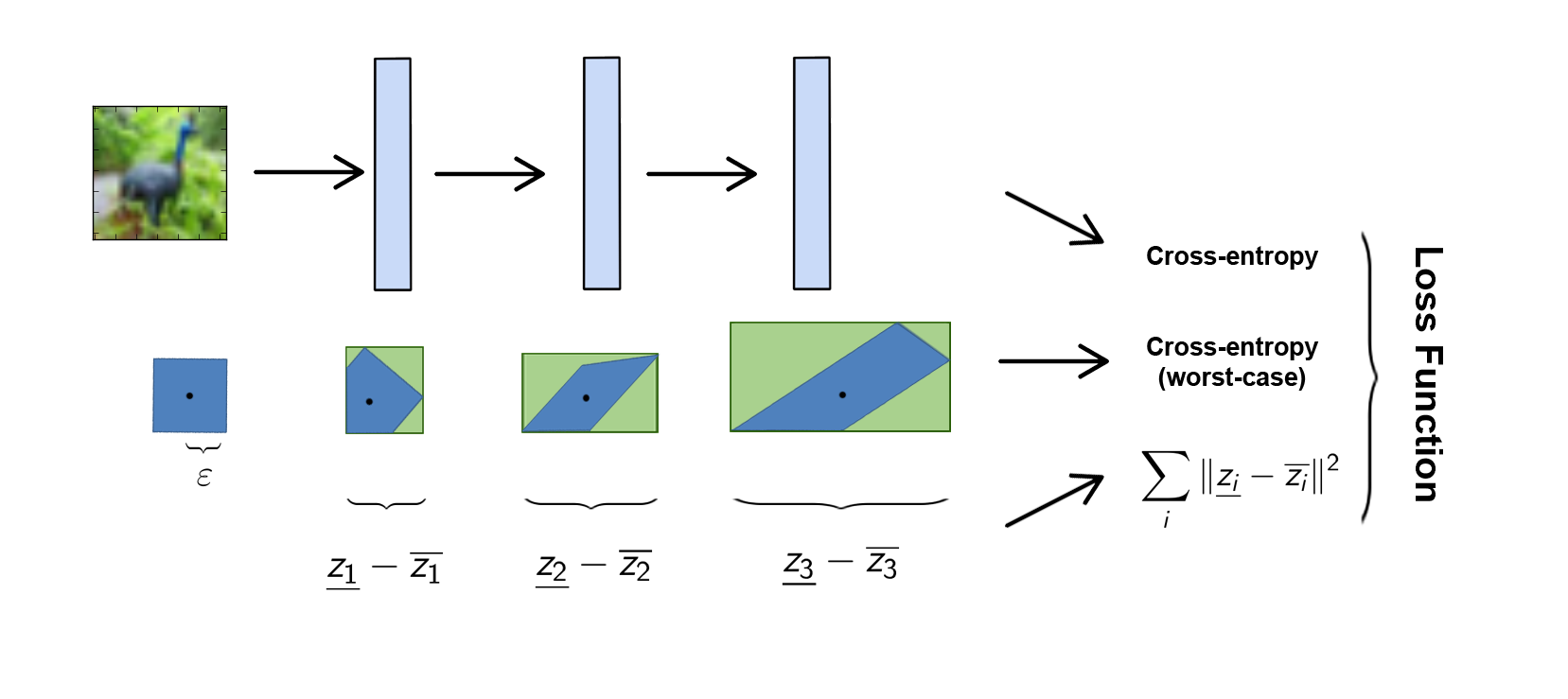}} 
\caption{The scheme of the proposed method. The original IBP loss is supplied with an additional term controlling the errors across layers.}
\label{fig:IBP_mse_scheme}
\end{figure}

In this contribution, we show that the training procedure of IBP can be significantly simplified, which results in more stable training and faster convergence. Our key idea relies on combining the IBP loss with an additional term, which controls the size of adversarial polytope across layers. More precisely, we minimize the size of outer bound of adversarial polytope at each layer jointly with the IBP cost function, see Figure \ref{fig:IBP_mse_scheme} for the illustration. As a result, our model is less sensitive to the change of the aforementioned IBP hyper-parameters, which makes it easier to use in practice.


Our contribution is the following:
\begin{enumerate}

\item We introduce a new term to the IBP loss function. Our modification allows to use larger perturbations at the initial stage of training and helps to stabilize the training.  Moreover, it requires a lower number of epochs to obtain comparable performance to IBP. In consequence, our model can be seen as a very efficient technique for constructing provable robust models, which can be applied to large networks.
The proposed idea is not limited to IBP and can be incorporated in other robust training methods, such as the convex-optimization-based approaches \cite{wong2017provable}, \cite{DualConvex}. It also helps to reduce hyperparameter tuning (particularly dynamics of $\epsilon$ during the training).

\item We give an insight on instability of IBP and show that this effect is correlated with a lack of minimization of interval bounds in hidden layers. Looking from a different perspective --- we observe that IBP (implicitly) minimizes the interval bounds in hidden layers when the training is convergent.

\item Conducted experiments support the research hypothesis, that the additional term in the loss function stabilizes the training, improves its efficiency and guides the network in the early stage of training. In the most challenging settings for the CIFAR-10, we are able to get better results (verified test error) even using much smaller network than the one used in the IBP's best performance. We show concrete examples, where IBP fails (or gets stuck in local minimum for a long time), whereas the new loss function allows to train the model in a stable fashion. 
\end{enumerate}

\section{Related work}

The study on adversarial examples has begun from \cite{szegedy2013intriguing}, in which the authors noticed that neural networks are fragile to targeted perturbations. Since then, numerous attacks and defenses have followed  \cite{papernot2016limitations, moosavi2017universal, xiao2018generating, kurakin2016adversarial, tramer2017ensemble, yuan2019adversarial}. One extremely effective way to defend against adversarial examples is to generate such examples in the training stage \cite{madry2017towards}. In this approach, we try to mimic the adversary and simulate his behaviour. While this strategy provides practical benefits, one cannot guarantee that other attacks does not exist. 

To construct provable defenses, we aim to produce certificates that no perturbation within a fixed norm can change a given class label. There is a number of works using exact solvers to verify
the robustness against adversarial attacks. These methods employ either integer programming approaches \cite{lomuscio2017approach, tjeng2018evaluating, dutta2018output, xiao2018training, cheng2017maximum}, or Satisfiability
Modulo Theories (SMT) solvers \cite{katz2017reluplex, carlini2017provably, ehlers2017formal}. A downside of these methods is their high computational complexity due to the NP-completeness of the problem. In consequence, vast majority of these methods do not scale well to large or even medium size networks.

To speed up the training of verifiably robust models, one can bound a set of activations reachable through a norm-bounded perturbation \cite{salman2019convex, liu2019certifying}. In \cite{wong2017provable}, linear programming was used to find the convex outer bound for ReLU networks. This approach was later extended to general non-ReLU neurons \cite{zhang2018efficient}. As an alternative, \cite{mirman2018differentiable, singh2018fast, singh2019abstract} adapted the framework of `abstract transformers' to compute an approximation to the adversarial polytope using the SGD training. This allowed to train the networks on entire regions of the input space at once. Interval bound propagation \cite{gowal2018effectiveness} applied the interval arithmetic to propagate axis-aligned bounding box from layer to layer. Analogical idea was used in \cite{dvijotham2018training}, in which the predictor and the verifier networks were trained simultaneously. While these methods are computationally appealing, they require careful tuning of hyper-parameters to provide tight bounds on the verification network. Finally, there are also hybrid methods, which combine exact and relaxed verifiers \cite{bunel2018unified, singh2018boosting}.

\section{Interval bound propagation}
\label{sec:IBP}
In this section we first give the main idea behind training provable robust models. Next, we recall the IBP framework based on the interval arithmetic. Finally, we present our model, which is the extension of the IBP approach.

\subsection{Training robust classifiers}

We consider a feed-forward neural network $f:\R^D \to \R^N$ designed for a classification task. The network is composed of $K$ layers given by $K$ transformations: 
$$
z_k = h_k(z_{k-1}), \text{ for } k=1,\ldots,K.
$$
In practice, $h_k$ is either an affine transformation or nonlinear monotonic function such as ReLU or Sigmoid. In the training stage, we feed the network with pairs of input vector $z_0 = x$ and its correct class label $y_{true}$ and minimize the cross-entropy with softmax applied to the output logits $z_K$.

In the adversarial attack, any test vector $x$ can be perturbed by some $\Delta$ with $l_\infty$ norm-bounded by $\epsilon$, for a small fixed $\epsilon > 0$. Thus the input to the network can by any point in $D$-dimensional hyper-cube: 
$$
I_\epsilon(x) = I(x - \epsilon, x + \epsilon) = [x_1-\epsilon, x_1 + \epsilon] \times \ldots \times [x_D-\epsilon, x_D + \epsilon].
$$
centered at $x$ with side length $2\epsilon$. This set is transformed by a neural network $f$ into some convex set called adversarial polytope: 
$$
\Z_\epsilon(x) = \{f(z): z \in I_\epsilon(x)\}.
$$
To design provable defense against adversarial attack, we have to ensure that class label $y_{true}$ does not change for any output $z_K \in \Z_\epsilon(x)$. In other words, all inputs from the hyper-cube $I_\epsilon(x)$ should be labeled as $y_{true}$ by a neural network $f$. In this context, a fraction of incorrectly classified examples on the test set is called the \textit{verified test error}.

\subsection{Verifiable robustness using IBP}

Exact verification of the model robustness may be difficult even for simple neural networks. Thus we usually look for an easier task computing loose outer bound of $\Z_\epsilon(x)$ and control the class label inside this bound. In the IBP approach \cite{gowal2018effectiveness}, we find the smallest bounding box at each layer that encloses the transformed bounding box from the previous layer. In other words, we bound the activation $z_k$ of each layer by an axis-aligned bounding box 
$$
I(\underline{z}_k, \overline{z}_k) = [\underline{z}_{k,1}, \overline{z}_{k,1}] \times \ldots \times [\underline{z}_{k,D_k}, \overline{z}_{k,D_k}].
$$

In the case of neural networks, finding such a bounding box from layer to layer fashion can be computed efficiently using the interval arithmetic. By applying the affine layer 
$$
h_k(z_{k-1}) = W_k z_{k-1} + b_k
$$ 
to $I(\underline{z}_{k-1}, \overline{z}_{k-1})$, the smallest bounding box $I(\underline{z}_{k}, \overline{z}_{k})$ for output $z_k$ is given by
$$
\begin{array}{l}
    \mu_{k-1} = \frac{\overline{z}_{k-1} + \underline{z}_{k-1}}{2},\\[0.8ex]
    r_{k-1} = \frac{\overline{z}_{k-1} - \underline{z}_{k-1}}{2},\\[0.8ex]
    \mu_k = W_k \mu_{k-1} + b_{k-1}\\[0.8ex]
    r_k = |W_k| r_{k-1}\\[0.8ex]
    \underline{z}_k = \mu_k - r_k,\\[0.8ex]
    \overline{z}_k = \mu_k + r_k,\\
\end{array}
$$
where $|\cdot|$ is an element-wise absolute value operator. For a monotonic activity function $h_k$, we get the interval bound defined by:
$$
\begin{array}{l}
      \underline{z}_k = h(\underline{z}_{k-1}),\\[0.8ex]
      \overline{z}_k = h(\overline{z}_{k-1}).
\end{array}
$$

To obtain a provable robustness in the classification context, we consider the worst-case prediction for the whole interval bound $I(\underline{z}_K, \overline{z}_K)$ of the final logits. More precisely, we need to ensure that the whole bounding box is classified correctly, i.e. no perturbation changes the correct class label. In consequence, the logit of the true class is equal to its lower bound and the other logits are equal to their upper bounds:
$$
\hat{z}_{K,y}(\epsilon) = \left\{
\begin{array}{ll}
     \overline{z}_{K,y}, & \text{ for } y \neq y_{true},\\[0.8ex]
     \underline{z}_{K,y_{true}}, & \text{ otherwise }.
\end{array}
\right.
$$
Finally, one can apply softmax with the cross-entropy loss to the  logit vectors $\hat{z}_K(\epsilon)$ representing the worst-case prediction.

As shown in \cite{gowal2018effectiveness}, computing interval bounds uses only two forward passes through the neural network, which makes this approach appealing from a practical perspective. Nevertheless, a direct application of the above procedure with a fixed $\epsilon$  may fail because propagated bounds are too loose especially for very deep networks (see also wrapping effect illustrated in Figure \ref{fig:wrapping_effect}) . To overcome this problem Gowal et al. supplied the above interval loss with a typical cross-entropy cost applied to original non-interval data:
$$
\IBP = \kappa \loss(z_K, y_{true}) + (1-\kappa) \loss(\hat{z}_k(\epsilon), y_{true}),
$$
where $\kappa$ is a trade-off parameter. In the initial training phase, the model uses only classical loss function applied to non-interval data ($\kappa = 1$). Next, the weight of the interval loss is gradually increased up to $\kappa = 1/2$. Moreover, the training starts with the small perturbation radius $\epsilon$, which is also increased in later epochs. The training process is sensitive to these hyperparameters and finding the correct schedule for every new data set can be problematic and requires extensive experimental studies. This makes the whole training procedure time consuming, which reduces practicality of this approach.

\subsection{Constrained interval bound propagation}

To make IBP less sensitive to the training settings and provide more training stability (particularly for bigger $\epsilon$), we propose to enhance the cost function. We want to directly control the bounding boxes at each layer of the network. More precisely, in addition to the IBP loss, we minimize the size of the outer interval bound at each layer. Thus our cost function equals 
$$
\our = \kappa \loss(z_K, y_{true}) + (1-\kappa) \loss(\hat{z}_k(\epsilon), y_{true}) + \sum_{k=1}^{K} \| \overline{z}_k - \underline{z}_k \|^2.
$$

We argue that such the addition would help to circumvent limitations of the original IBP.
First, gradients would be calculated not only with respect to the the last layer but to all hidden layers. This should bring more training stability, especially at the early training stage. Second, we expect it would be easier for a model to have small interval bounds in the final layer when bounds are constrained in hidden layers. And indeed our experimental results support these research hypotheses. 

\section{Experiments}

Here we report our experiments, which show the effect of the proposed loss function and give some insight why it is beneficial to minimize the interval bounds in hidden layers. We implement our ideas with PyTorch library \cite{Pytorch} and for a fair comparison we reimplement the original IBP using the same framework. We conduct the experiments on three datasets, namely CIFAR-10 \cite{cifar10}, SVHN \cite{svhn} and MNIST \cite{mnist}. The neural network architectures used in the experiments are the same as in \cite{gowal2018effectiveness} and these are 3 convolutional nets called \textit{small}, \textit{medium} and \textit{large}, see Table \ref{tab:arch} for details. In all experiments adversarial perturbations are within $\ell_{\infty}$ norm-bounded ball. If not stated otherwise, we always apply the original training procedure and hyper-parameters used in \cite{gowal2018effectiveness}. For CIFAR-10 and SVHN, we use data augmentation (adding random translations and flips, normalizing each image channel using the channel statistics from the training set).

\begin{table}[t]
\small
\caption{Architectures used in the experiments. There are 4 parameters for the convolutional layer (Conv2d): a number of input and output filters and a size of a filter and a stride. For the fully connected layer the parameter denotes a number of outputs.\label{tab:arch}}
\begin{tabular}{l|l|l}

\hline
small                            & medium                           & large                            \\ \hline
Conv2d(input\_filters, 16, 4, 2) & Conv2d(input\_filters, 32, 3, 1) & Conv2d(input\_filters, 64, 3, 1) \\
Conv2d(16, 32, 4, 1)             & Conv2d(32, 32, 4, 2)             & Conv2d(64, 64, 3, 1)             \\
Fully\_connected(100)            & Conv2d(32, 64, 3, 1)             & Conv2d(64, 128, 3, 2)            \\
Fully\_connected(10)             & Conv2d(64, 64, 4, 2)             & Conv2d(128, 128, 3, 1)           \\
                                 & Fully\_connected(512)            & Conv2d(128, 128, 3, 1)           \\
                                 & Fully\_connected(512)            & Fully\_connected(200)            \\
                                 & Fully\_connected(10)             & Fully\_connected(10)            
\end{tabular}
\end{table}

\subsection{Faster convergence}

\begin{figure}[t] 
\center{\includegraphics[width=\textwidth] {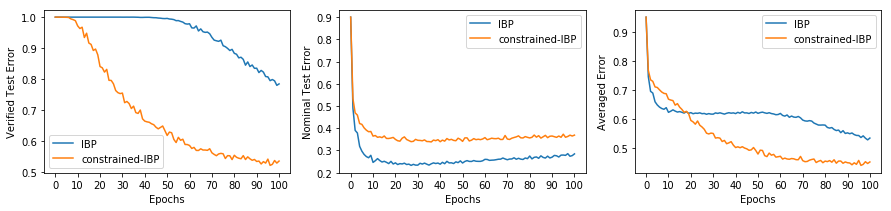}} 
\center{\includegraphics[width=\textwidth] {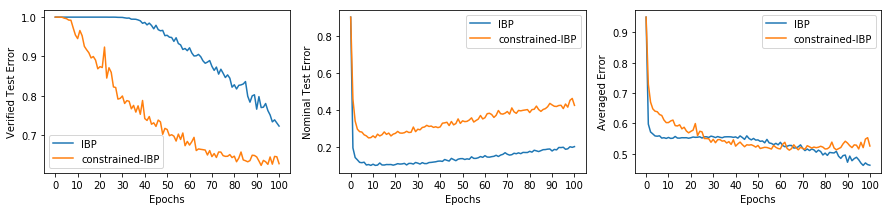}} 
\caption{Verifiably robust training for the CIFAR-10 (top row) and SVHN (botton row) with adversarial perturbations bounded by $\epsilon=8/255$. We report the verified test error, test error and the average between these two metrics. Low nominal test errors are reached very quickly because the training starts basically as the training of a regular classifier and and the robust term is introduced gradually (as explained in Section \ref{sec:IBP}). 
}
\label{fig:CIFAR_faster_train}
\end{figure}

First, we highlight that our approach minimizes the verified test error much faster than IBP. Since the performance of both methods on MNIST is comparable, we only report the results on most challenging cases of CIFAR-10 and SVHN with maximal perturbation radius $\epsilon=8/255$.

It is evident from Figure \ref{fig:CIFAR_faster_train} that the difference between both methods is substantial. In the case of CIFAR-10 after 100 epochs, the verified test error is over 20 percentage points lower, whereas the nominal error is close. The shape of the curves for SVHN is similar, but the gain in verified accuracy is slightly lower; after 50 epochs the verified error of $\our{}$ is also 20 percentage points lower than the one obtained by IBP, while after 100 epochs the difference is around 10 percentage points.

We stress that we took the training schedule directly from the IBP method (in particular changes in $\epsilon$ and $\kappa$), which were tuned for that method. Selecting optimal parameters for $\our{}$ opens the possibility for even better results and better trade-offs between the verified and the nominal test errors. In the next section, we investigate the  more dynamic $\epsilon$ changes, which would make the training even faster.




\subsection{More stable training}

Gowal et al. stated that their method needs to slowly increase $\epsilon$ (from 0 to $\epsilon_{train}$) to provide stability and convergence during the training. For example, for CIFAR-10, this `ramp-up' phase lasts 150 epochs. It raises a natural question whether we could speed-up the $\epsilon$ increase and whether our new term in the loss function is helpful in this regard.

\begin{figure}[t] 
\center{\includegraphics[width=\textwidth] {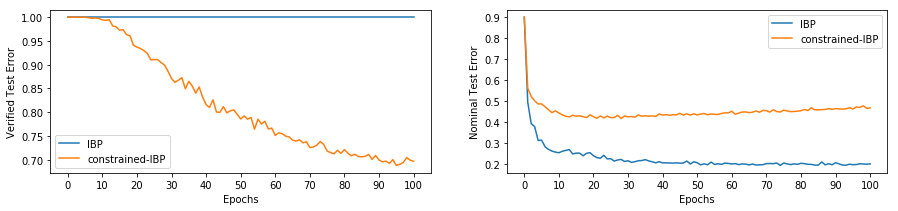}} 
\caption{Verifiably robust training for the CIFAR-10 dataset with adversarial perturbations bounded by $\epsilon=4/255$. Experiments done on the large architecture.}
\label{fig:CIFAR_stable_train}
\end{figure}

First we show that even if we keep the original $\epsilon$ changes, IBP may stuck in a local minimum for a very long time. The experiment was done on CIFAR-10 with the large architecture and $\epsilon=4/255$. The test error goes down very quickly, reaching 0.2, whereas the verified test error remains 100\% for over 100 epochs, see Figure \ref{fig:CIFAR_stable_train}. On the contrary, our approach steadily minimizes the verified test error. 

\begin{figure}[t] 
\center{\includegraphics[width=\textwidth] {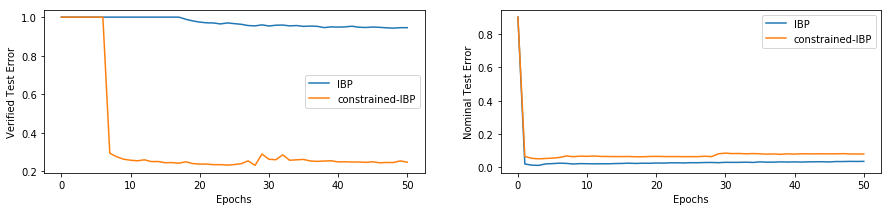}} 
\caption{Verifiably robust training for the MNIST dataset with adversarial perturbations bounded by $\epsilon=0.4$. Experiments done on the small architecture. Perturbation radius $\epsilon$ was increased 2.5 times faster than in \cite{gowal2018effectiveness}.}
\label{fig:MNIST_stable_train}
\end{figure}

Next, we investigate the more dynamic $\epsilon$ changes to reduce the training time. For the MNIST dataset, increasing $\epsilon$ 2.5 faster results in lack of convergence for the original IBP method, see Figure \ref{fig:MNIST_stable_train}. On the other hand, the additional term in the loss function helps to stabilize the training and obtain the minimization of verified error. 

\subsection{Minimizing interval bounds in hidden layers is desired}

\begin{figure}[t] 
\center{\includegraphics[width=\textwidth] {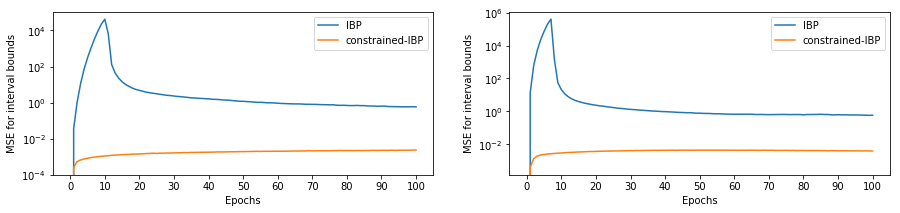}} 
\caption{Verifiably robust training for CIFAR-10 (left) and SVHN (right) with $\epsilon=8/255$. We report the total sum of mean squared errors between lower and upper interval bounds in hidden layers. }
\label{fig:bounds_stable}
\end{figure}

In our approach, we add the additional term to the loss function which helps to minimize the interval bounds in hidden layers. Interestingly, the IBP method also tries to implicitly keep the interval bounds stable in the hidden layers, see top row of Figure \ref{fig:bounds_stable}. In fact this is a natural behaviour, because it is easier for a model to have small interval bounds in the final layer when bounds are possibly small in hidden layers. Nevertheless, the additional term in our approach stabilizes bounds faster and their values are a few orders of magnitude lower. 

\begin{figure}[t] 
\center{\includegraphics[width=0.37\textwidth] {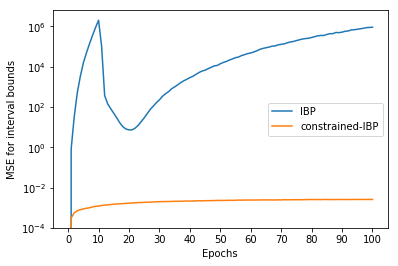}} 
\caption{Verifiably robust training for MNIST with more dynamic $\epsilon$ changes. We report the total sum of mean squared errors between lower and upper interval bounds in hidden layers. }
\label{fig:bounds_stable2}
\end{figure}

To gain more insight, we also checked what happens to the interval bounds during unstable training, such as the one shown in Figure \ref{fig:MNIST_stable_train} for MNIST. It turns out that in such settings (more dynamic $\epsilon$ changes), the IBP method is unable to stabilize the interval bounds in hidden layers, see Figure \ref{fig:bounds_stable2}. This observation supports our hypothesis that verified error is highly correlated with the interval bounds in the hidden layer. In consequence, it is beneficial to explicitly encourage the model to keep the interval bounds low across layers. 


\subsection{Comparison with the original Interval Bound Propagation}

For the sake of completeness we provide the comparison between IBP and our approach in terms of final error rates. We report the test error, the error rate under the PGD attack and the verified bound on the error rate. All these numbers are obtained after the complete training. We stress that all hyperparameters (particularly schedules of $\epsilon$, $\kappa$ and learning rate) were left the same as in \cite{gowal2018effectiveness}. 

It is evident from Table \ref{tab:ben} that without any hyperparameters tuning, our approach gives comparable results to IBP. However, the reported levels of errors are reached faster and the training is done in more stable way --- our main points described in the previous sections. We highlight the results for CIFAR-10 with challenging $\epsilon=8/255$, where our modification lets even the smallest model beat the verified error obtained by Gowal et al for this dataset. For the largest network, we think the original hyperparameters (number of epochs, learning rate schedule) cause overfitting for our method, i.e. the model learns quickly and spends most of the training time on fitting to the train set.

\begin{table}[h]
\centering
\begin{tabular}{lc|llll}
\hline
Dataset/model & \multicolumn{1}{l|}{Epsilon} & Method & Test error & PGD & Verified \\ \hline

            &    & Nominal & 20.06\% & 95.76\%  & 100\%  \\
CIFAR-10/small & 8/255 & {IBP (Gowal et al.)} & {39.33\%} &\textbf{52.22\%} & {63.58\%} \\
             &       & \cellcolor[HTML]{ECF4FF}constrained-IBP & \cellcolor[HTML]{ECF4FF}47.22\%  & \cellcolor[HTML]{ECF4FF}{56.14\%} & \cellcolor[HTML]{ECF4FF}\textbf{60.12\%} \\ \hline
            &  & Nominal & 13.97\% & 89.59\% & 100\% \\
CIFAR-10/medium & 8/255 & IBP (Gowal et al.) & 18.88\%  & \textbf{48.32\%}  & 100\%  \\
            &  & \cellcolor[HTML]{ECF4FF}constrained-IBP & \cellcolor[HTML]{ECF4FF}{49.29\%} & \cellcolor[HTML]{ECF4FF}{57.69\%}& \cellcolor[HTML]{ECF4FF}\textbf{61.81\%} \\ \hline
     &       & Nominal &  12.95\% & \textbf{50.24\%} & 100\%  \\
CIFAR-10/large & 8/255 & IBP (Gowal et al.) & 40.55\% & 56.65\% & 65.89\%  \\
     &      & \cellcolor[HTML]{ECF4FF}constrained-IBP & \cellcolor[HTML]{ECF4FF}{52.04\%}& \cellcolor[HTML]{ECF4FF}{59.89\%} & \cellcolor[HTML]{ECF4FF}\textbf{63.89\%}\\ \hline \hline

    &   & Nominal & 0.65\%  & 99.63\%   & 100\% \\
MNIST/small & 0.4 & {IBP (Gowal et al.)} & {2.62\%} & \textbf{14.14\%} & \textbf{20.74\%} \\
    &   & \cellcolor[HTML]{ECF4FF}constrained-IBP & \cellcolor[HTML]{ECF4FF}{7.24\%} & \cellcolor[HTML]{ECF4FF}{17.24\%} & \cellcolor[HTML]{ECF4FF}{21.31\%} \\ \hline
    &   & Nominal & 1.06\% & 99.64\%  & 100\% \\
MNIST/medium & 0.4 & IBP (Gowal et al.) & 1.66\% & 12.16\% & 17.5\% \\
    &   & \cellcolor[HTML]{ECF4FF}constrained-IBP & \cellcolor[HTML]{ECF4FF}{1.91\%}& \cellcolor[HTML]{ECF4FF}\textbf{9.7\%}& \cellcolor[HTML]{ECF4FF}\textbf{16.53\%}\\ \hline
    &   & Nominal & 0.65\%  & 99.64\%   & 100\% \\
MNIST/large & 0.4 & IBP (Gowal et al.) &1.66\%  &10.34\%  &\textbf{15.01\%}  \\
    &   & \cellcolor[HTML]{ECF4FF}constrained-IBP & \cellcolor[HTML]{ECF4FF}{1.62\%}& \cellcolor[HTML]{ECF4FF}\textbf{8.32\%}& \cellcolor[HTML]{ECF4FF}{15.58\%} \\ \hline \hline

    &   & Nominal & 9.05\% & 82.05\%  & 100\%  \\
SVHN/small & 8/255 & {IBP (Gowal et al.)} &  26.60\% &  \textbf{48.50\%} &  60.87\% \\
&   & \cellcolor[HTML]{ECF4FF}constrained-IBP & \cellcolor[HTML]{ECF4FF}{36.10\%} & \cellcolor[HTML]{ECF4FF}{49.31\%} & \cellcolor[HTML]{ECF4FF}\textbf{53.97\%} \\ \hline
    &   & Nominal & 5.84\% & 65.60\% & 100\%  \\
SVHN/medium & 8/255 & IBP (Gowal et al.) & 36.58\% & 48.79\% & 55.95\% \\
    &   & \cellcolor[HTML]{ECF4FF}constrained-IBP & \cellcolor[HTML]{ECF4FF}{21.55\%} & \cellcolor[HTML]{ECF4FF}\textbf{39.23\%} & \cellcolor[HTML]{ECF4FF}\textbf{54.32\%} \\ \hline
    &   & Nominal & 5.78\% & 77.25\% & 100\% \\
SVHN/large & 8/255 & IBP (Gowal et al.) & 20.00\% &\textbf{37.06\%}  &\textbf{52.37\%}  \\
    &   & \cellcolor[HTML]{ECF4FF}constrained-IBP & \cellcolor[HTML]{ECF4FF}{42.62\%} & \cellcolor[HTML]{ECF4FF}{53.90\%} & \cellcolor[HTML]{ECF4FF}{59.16\%} \\ \hline

\end{tabular}
\caption{Comparison between IBP by Gowal et al. and our modification. For the point of reference, we also report numbers for the regular training (Nominal), without adversarial perturbations. The PGD error rate is calculated using 200 iterations of PGD and 10 random restarts. \label{tab:ben}}
\end{table}

\section{Conclusion}

Most techniques for training verifiably robust classifiers are computationally demanding. In this paper, we used a simple but promising technique based on the interval arithmetic \cite{gowal2018effectiveness}, which needs only two standard network passes to process the input perturbed within the norm-bounded ball. Although a single iteration can be performed fast, the whole training process requires careful tuning and many epochs to converge. As a remedy, we proposed to additionally minimize the size of an outer bound of the adversarial polytope across hidden layers. This modification was motivated by the observation that IBP implicitly minimize these bounds in the case of the successful, convergent training. By adding this constraint explicitly, the model become less sensitive to the change of hyper-parameters and, in consequence, we could increase the perturbation radius more dynamically to the desired value, which makes the training faster. As a result, we were able to obtain the lowest verified error in the most challenging case of CIFAR-10 with the perturbation radius $8/255$ using only the small architecture.


\newpage

\bibliographystyle{plainnat}
\bibliography{ref}

\end{document}